\documentclass[letterpaper, 10 pt, conference]{ieeeconf}
\IEEEoverridecommandlockouts
\usepackage{xargs} 
\usepackage{graphicx}
\usepackage{color,soul}
\usepackage{pgfgantt}
\usepackage{pdflscape}
\usepackage{lipsum}
\usepackage{pstricks}
\usepackage{import}
\usepackage{color}
\usepackage{xcolor}
\definecolor{midgrey}{RGB}{100, 100, 100}
\definecolor{darkgreen}{RGB}{0, 100, 50}
\usepackage{layouts}
\usepackage{booktabs}

\usepackage{multirow}
\usepackage{siunitx}
\usepackage{amsmath,amssymb}
\DeclareMathOperator{\E}{\mathbb{E}}
\DeclareMathOperator*{\argmax}{argmax}

\DeclareFontFamily{U}{mathc}{}
\DeclareFontShape{U}{mathc}{m}{it}%
{<->s*[1.03] mathc10}{}

\DeclareMathAlphabet{\mathscr}{U}{mathc}{m}{it}

\usepackage{algorithm}
\usepackage[noend]{algpseudocode}
\usepackage{algpseudocode}
\usepackage{stackengine}

\usepackage{pgf}
\usepackage[acronym]{glossaries}
\newacronym{RL}{RL}{reinforcement rearning}

\usepackage[symbol]{footmisc}

\makeatletter
\def\input@path{{./figures/}}
\makeatother

\makeatletter
\edef\@figdir{_pgfcache}

\catcode`\#=11
\catcode`\|=6
\newcommand{\@basicpgfpreamble}[1]{%
    \unexpanded{%
        \documentclass{standalone}^^J
        \usepackage{pgf}^^J
        \RequirePackage{luatex85}^^J
        \let\oldpgfimage\pgfimage^^J
        \renewcommand{\pgfimage}[2][]{\oldpgfimage[#1]{|1/#2}}^^J
    }%
}
\catcode`\#=6
\catcode`\|=12

\let\@pgfpreamble\@basicpgfpreamble

\newcommand{\setpgfpreamble}[1]{%
    \renewcommand{\@pgfpreamble}[1]{\@basicpgfpreamble{##1}\unexpanded{#1}}
}

\newcounter{@pgfcounter}
\newwrite\@pgfout
\newread\@pgfin

\newcommand{\importpgf}[3][]{%
    \IfFileExists{#2/#3}{}{\errmessage{importpgf: File #2/#3 not found}}%
    \edef\@figfile{\jobname-\the@pgfcounter}%
    \providecommand{\@writetempfile}{}%
    \renewcommand{\@writetempfile}[1]%
    {%
        \immediate\openout\@pgfout=##1%
        \immediate\write\@pgfout{\@pgfpreamble{#2}}%
        \immediate\write\@pgfout{\string\begin{document}}%
        \immediate\openin\@pgfin=#2/#3%
        \begingroup\endlinechar=-1%
            \loop\unless\ifeof\@pgfin%
                \readline\@pgfin to \@fileline%
                \ifx\@fileline\@empty\else%
                    \immediate\write\@pgfout{\@fileline}%
                \fi%
            \repeat%
        \endgroup%
        \immediate\closein\@pgfin%
        \immediate\write\@pgfout{\string\end{document}}%
        \immediate\closeout\@pgfout%
    }%
    \def\@compile%
    {%
        \immediate\write18{lualatex -interaction=batchmode -output-directory="\@figdir" \@figdir/\@figfile.tex}%
    }%
    \IfFileExists{\@figdir/\@figfile.pdf}%
    {%
        \@writetempfile{\@figdir/tmp.tex}%
        \edef\@hashold{\pdfmdfivesum file {\@figdir/\@figfile.tex}}%
        \edef\@hashnew{\pdfmdfivesum file {\@figdir/tmp.tex}}%
        \ifnum\pdfstrcmp{\@hashold}{\@hashnew}=0%
            \relax%
        \else%
            \@writetempfile{\@figdir/\@figfile.tex}%
            \@compile%
        \fi%
    }%
    {%
        \@writetempfile{\@figdir/\@figfile.tex}%
        \@compile%
    }%
    \IfFileExists{\@figdir/\@figfile.pdf}%
    {\includegraphics[#1]{\@figdir/\@figfile.pdf}}%
    {\errmessage{Error during compilation of figure #2/#3}}%
    \stepcounter{@pgfcounter}%
}
\makeatother

\usepackage[colorinlistoftodos,prependcaption,textsize=small]{todonotes}
\newcommandx{\unsure}[2][1=]{\todo[inline,linecolor=red,backgroundcolor=red!25,bordercolor=red,#1]{#2}}
\newcommandx{\change}[2][1=]{\todo[inline,linecolor=blue,backgroundcolor=blue!25,bordercolor=blue,#1]{#2}}
\newcommandx{\info}[2][1=]{\todo[inline,linecolor=OliveGreen,backgroundcolor=OliveGreen!25,bordercolor=OliveGreen,#1]{#2}}
\newcommandx{\improvement}[2][1=]{\todo[inline,linecolor=Plum,backgroundcolor=Plum!25,bordercolor=Plum,#1]{#2}}
\newcommandx{\thiswillnotshow}[2][1=]{\todo[disable,#1]{#2}}
\newcommand{\rework}[1]{%
	\todo[color=yellow,inline, caption={}]{\normalsize Rework: {#1}}%
}

\graphicspath{{./pictures/}{./figures/}}

\newcommand{\figurepath}{plots}
\newcommand{\figurepathnew}{./figures/}

\newcommand{\tablepath}{./tables}

\newcommand{\Astar}{$\text{A}^* $ }

\usepackage{scalerel,stackengine,amsmath}
\newcommand\equalhat{\mathrel{\stackon[1.5pt]{=}{\stretchto{%
				\scalerel*[\widthof{=}]{\wedge}{\rule{1ex}{3ex}}}{0.5ex}}}}

\usepackage[style=ieee,	sorting=none,bibencoding=utf8, maxcitenames=2, mincitenames=1, minbibnames=2, maxbibnames=2, backend=biber, citestyle=numeric-comp, isbn=false,doi=false, url=false, natbib=true]{biblatex}

\AtEveryBibitem{%
	\ifentrytype{book}{
		\clearfield{url}%
		\clearfield{urldate}%
		\clearfield{review}
		\clearfield{series}
		\clearfield{note}
		\clearlist{location}
		\clearname{editor}
		\clearfield{pages}
	}{}
	\ifentrytype{collection}{
		\clearfield{url}%
		\clearfield{urldate}%
		\clearfield{review}%
		\clearfield{series}
		\clearfield{note}
		\clearlist{location}
		\clearname{editor}
		\clearfield{pages}
	}{}
	\ifentrytype{incollection}{
		\clearfield{url}%
		\clearfield{urldate}%
	\clearfield{review}%
		\clearfield{series}
		\clearfield{note}
		\clearlist{location}
		\clearfield{pages}
		\clearname{editor}
	}{}
		\ifentrytype{inproceedings}{
		\clearlist{location}
		\clearfield{url}%
		\clearfield{urldate}%
		\clearfield{review}%
		\clearfield{series}
		\clearfield{publisher}
		\clearfield{note}	
		\clearfield{editor}
		\clearfield{pages}
		\clearname{editor}
	}{}
	\ifentrytype{article}{
		\clearlist{location}
		\clearfield{url}%
		\clearfield{urldate}%
		\clearfield{review}%
		\clearfield{series}
		\clearfield{publisher}
		\clearfield{note}	
		\clearfield{pages}
		\clearname{editor}
	}{}
}

\DeclareNameAlias{sortname}{last-first}
\DeclareNameAlias{default}{last-first}

 \usepackage[caption=false,font=footnotesize]{subfig}

 \usepackage{dblfloatfix}
\usepackage[colorinlistoftodos]{todonotes}        
\usepackage{setspace}
\usepackage{acronym}

\newcommand{\add}[2][]{\todo[color=blue!40,#1]{#2}{}}
\newcommand\optional[2][]{\todo[inline, color=cyan!40, caption={2do},
	#1]{\begin{minipage}{\textwidth-4pt}#2\end{minipage}}{}}

\newcommand{\set}[1]{\boldsymbol{#1}}
\renewcommand{\vec}[1]{\mathbf{#1}}

\graphicspath{{pics/}}

\newcommand\copyrighttext{%
	\scriptsize \textcolor{blue}{\textcopyright 2019 IEEE. Personal use of this material is permitted.  Permission from IEEE must be obtained for all other uses, in any current or future media, including reprinting/republishing this material for advertising or promotional purposes, creating new collective works, for resale or redistribution to servers or lists, or reuse of any copyrighted component of this work in other works}}
\newcommand\copyrightnotice{%
	\begin{tikzpicture}[remember picture,overlay]
	\node[anchor=north,yshift=-7.5pt] at (current page.north) {\fbox{\parbox{\dimexpr\textwidth-\fboxsep-\fboxrule\relax}{\copyrighttext}}};
	\end{tikzpicture}%
}

\begin{document}
%
\title{\begin{center} \LARGE \bf Addressing Inherent Uncertainty:  Risk-Sensitive Behavior Generation for Automated Driving using Distributional Reinforcement Learning \end{center}}

%
%
%

%
\author{Julian Bernhard$^{1}$, Stefan Pollok$^{1}$ and Alois Knoll$^{2}$%
	\thanks{$^{1}$Julian Bernhard and Stefan Pollok are with fortiss GmbH, An-Institut Technische Universit\"{a}t M\"{u}nchen, Munich, Germany}%
	\thanks{$^{2}$Alois Knoll is with Chair of Robotics, Artificial Intelligence and Real-time Systems, Technische Universit\"{a}t M\"{u}nchen, Munich, Germany}
}
%

\vspace{-4mm}
\maketitle

\copyrightnotice
\thispagestyle{empty}
\pagestyle{empty}

\global\csname @topnum\endcsname 0
\global\csname @botnum\endcsname 0


\begin{abstract} For highly automated driving above SAE level~3, behavior generation algorithms must reliably consider the inherent uncertainties of the traffic environment, e.g. arising from the variety of human driving styles. Such uncertainties can generate ambiguous decisions, requiring the algorithm to appropriately balance low-probability hazardous events, e.g. collisions, and high-probability beneficial events, e.g. quickly crossing the intersection. State-of-the-art behavior generation algorithms lack a distributional treatment of decision outcome. This impedes a proper risk evaluation in ambiguous situations, often encouraging either unsafe or conservative behavior. Thus, we propose a two-step approach for risk-sensitive behavior generation combining offline distribution learning with online risk assessment. Specifically, we first learn an optimal policy in an uncertain environment with Deep Distributional Reinforcement Learning. During execution, the optimal risk-sensitive action is selected by applying established risk criteria, such as the Conditional Value at Risk, to the learned state-action return distributions. In intersection crossing scenarios, we evaluate different risk criteria and demonstrate that our approach increases safety, while maintaining an active driving style. Our approach shall encourage further studies about the benefits of risk-sensitive approaches for self-driving vehicles.
\end{abstract}

\section{Introduction}
For highly automated driving above SAE level 3, behavior generation algorithms must reliably consider the inherent uncertainties of the traffic environment, e.g. arising from the variety of driving styles of other participants. Such uncertainties can generate ambiguous decisions, requiring the algorithm to appropriately balance low-probability hazardous events, e.g. collisions, and high-probability beneficial events, e.g. quickly crossing the intersection. A single, numeric value measuring the outcome of a decision does not appropriately characterize such an ambiguous situation, since it neglects the probability of events. Instead of using an expectation-based utility measure, humans resolve ambiguity by minimizing an adequate risk-metric over an outcome distribution~\cite{majumdar_how_2017}. Such risk-metrics better evaluate the potential harm of an action with respect to the probability of occurence. 

However, state-of-the-art behavior generation algorithms still lack a distributional treatment of risk. On the one hand, frequently used problem definitions for behavior generation, e.g. MDPs\footnote[2]{\label{fn:problem_defs}MDP: Markov Decision Process, POMDP: Partially-Observable MDP, MAMDP: Multi-Agent MDP} \cite{mirchevska_high-level_2018, bouton_reinforcement_2018}, POMDPs\footref{fn:problem_defs} \cite{hubmann_belief_2018, bouton_belief_2017} or MAMDPs\footref{fn:problem_defs} \cite{kurzer_decentralized_2018, lenz_tactical_2016,burger_cooperative_2018}, adhere to expectation-based return calculation as it is the conventional definition of optimality for such problems. On the other hand, most problem solvers, e.g. Deep Q-Learning~\cite{mirchevska_high-level_2018,isele_navigating_2018,isele_navigating_2017} and Monte Carlo Tree Search~\mbox{\cite{kurzer_decentralized_2018, lenz_tactical_2016}} for MDPs or Adaptive Belief Tree \cite{hubmann_belief_2018} for POMDPs, output the expected return instead of the return distribution.  

\begin{figure}[t]																																															
	\def\svgwidth{\columnwidth}
	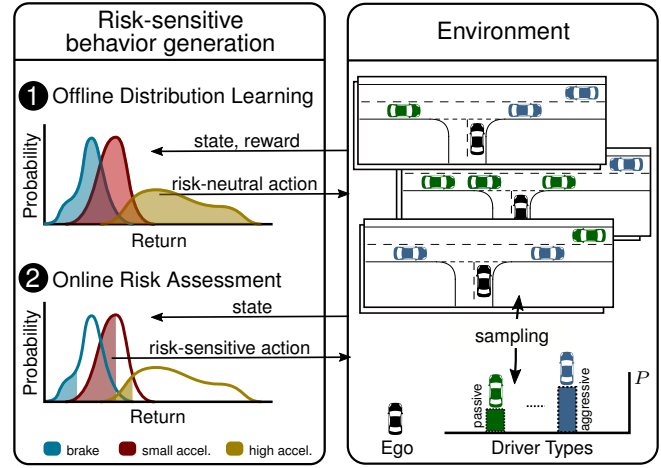 \vspace{-5mm}
	\caption{Our approach deals with an unknown \emph{episode-specific} driver type of a participant sampled from a known \emph{environment-specific} driver type distribution (right). Risk-neutral, state-action return distributions are trained offline for this environment and evaluated online regarding collision risk (left). This two-step risk-sensitive behavior generation approach increases safety in the face of behavioral uncertainty.} \label{fig:top_level}
	\vspace{-6mm}
\end{figure}

Interestingly, recent variants of Deep Reinforcement Learning enable learning of state-action return distributions~\cite{bellemare_distributional_2017, dabney_distributional_2017, dabney_implicit_2018}, motivating an approach for risk-sensitive behavior generation. Our two-step approach combines offline learning of the return distribution with online risk assessment (\mbox{Fig. \ref{fig:top_level}}). It demonstrates the advantages of using risk-sensitive return metrics to increase safety in the face of behavioral uncertainty. 
Specifically, we use Deep Distributional Q-Learning to learn the risk-neutral, state-action return distributions in environments with an unknown \emph{episode-specific} behavior type of a participant sampled from a known \emph{environment-specific} behavior type distribution. During execution, the optimal action is then selected based on a distortion risk metric applied to the learned state-action return distributions. 
																																					
The main contributions of this work are:
\begin{itemize}
	\item A risk-sensitive behavior generation approach combining offline Deep Distributional Reinforcement Learning with online risk assessment.  
	\item A benchmark of continuous observation spaces suitable for Deep~Q-Learning in intersection scenarios.
	\item An evaluation of risk metrics applicable for behavior generation of autonomous vehicles.
	\item A demonstration of the safety benefits of risk-sensitive behavior generation in environments with behavioral uncertainty.
\end{itemize}

This work is structured as follows: First, we present related work and introduce our approach. Next, we present the experiment setup with a benchmark of neural network observation spaces, followed by a qualitative and quantitative evaluation of our method.

\section{Related Work} \label{sec:related_work}
A behavior generation algorithm should consider the interactions between participants to successfully navigate in congested traffic, e.g. crowded intersections. 
Different variants exist to model interactive human behavior.

\emph{Cooperative approaches} assume that all agents optimize a global cost function. Solving this multi-agent MDP either with optimization- \cite{sadigh_planning_2016, burger_cooperative_2018} or search-based methods \cite{lenz_tactical_2016, kurzer_decentralized_2018-1} yields a globally optimal solution defining also the ego agent's behavior. However, the equilibrium assumption neglects the uncertainty inherent to human interactions. \emph{Probabilistic approaches} model the uncertainty about the behavior of other participants as hidden state in a POMDP and solve the problem mainly using sampling-based approaches either offline \cite{bouton_belief_2017} or online \mbox{\cite{hubmann_belief_2018, bai_intention-aware_2015}}. Both, cooperative and probabilistic approaches, face the problem of a combinatorially increasing number of maneuvering options with a growing number of participants. To achieve real-time capability, these algorithms limit the planning horizon \cite{hubmann_belief_2018}, consider only interactions with the nearest participants~\cite{lenz_tactical_2016, burger_cooperative_2018} and apply computationally simple traffic prediction models \cite{lenz_tactical_2016, hubmann_belief_2018, }, e.g. the Intelligent Driver Model. \mbox{\citet{menendez-romero_courtesy_2018}} define a \emph{probabilistic cooperative approach} for highway merging. However, their approach assumes a discrete formulation of the participants' intentions.  

\emph{Deep Reinforcement Learning} (DRL) promises interaction-aware decision making at lower computational cost. It learns the expected return of an action by interacting with other participants in simulation. During online planning, the agent exploits this experience.  \citet{isele_navigating_2017} apply Deep Q-Networks (DQN) to intersection crossing and extend it to occlusion handling in~\cite{isele_navigating_2018}. \mbox{\citet{wolf_adaptive_2018}} evaluate semantic state space definitions for DQNs in highway scenarios. Both approaches apply deterministic traffic models. Yet, even with deterministic models frequently a small percentage of collisions remains. This \emph{epistemic or parametric uncertainty} arises from imperfect information of the learning algorithm about the problem \cite{dilokthanakul_deep_2018}, e.g. coming from insufficient exploration or inexact minimization of the loss function. To overcome epistemic uncertainty when using reinforcement learning for autonomous driving behavior generation, one can combine DRL with a search process to allow escaping from local optima of the learned policy \cite{bernhard_experience-based_2018, paxton_combining_2017} or add an additional safety layer to avoid insecure actions \cite{mirchevska_high-level_2018,mukadam_tactical_2017, shalev-shwartz_safe_2016}.

In contrast to epistemic uncertainty, our work deals with \emph{inherent or aleatoric uncertainty} in the environment, e.g. arising from uncertainty about the behavior of other participants. To avoid unsafe decisions in such domains, risk-sensitive reinforcement learning employs an optimization criterion balancing the return and the risk of an action \cite{garcia_comprehensive_2015}. Such risk criteria and their application to the field of robotics are discussed in \cite{majumdar_how_2017}. \citet{dabney_implicit_2018} combine a novel, non-parametric approach for return distribution estimation using Deep Distributional Reinforcement Learning (DDRL) with risk-sensitive action selection. They outperform previous DDRL approaches in the domain of Atari games. The risk preferences of humans in driving scenarios are evaluated in~\cite{majumdar_risk-sensitive_2017} using Inverse Reinforcement Learning. To deal with behavioral uncertainty in DRL, \citet{bouton_reinforcement_2018} add safety rules that block actions when they violate a safety constraint with certain probability. The probabilistic safety measure is calculated separately for each participant in a discretized state space. At an intersection with two participants their approach yields zero collisions. However, the approach neglects interactions between other participants and does not scale efficiently to more complex scenarios.

To the best of our knowledge, our work is the first which addresses the inherent uncertainty of traffic environments with risk-sensitive optimization criteria. Our algorithm avoids a rule-based formulation of safety or discretization of the state space. It is solely based on the reward definition and easily interpretable risk evaluation metrics. Further, we demonstrate the advantages of Distributional Reinforcement Learning for autonomous vehicle behavior generation.

\section{Problem Definition}
There exist various types of inherent uncertainties. 
We focus on the inherent uncertainty that arises from the interaction with other traffic participants of varying driving styles. We formulate the problem as Stochastic Bayesian Game~(SBG)~\cite{albrecht_belief_2016}. The SBG models other agents' behaviors based on a behavior type space and distribution. We can adopt this notion to our domain: A behavior type corresponds to a human driving style, e.g. "aggressive" or "passive". The type distribution models the occurrence frequencies of the driving styles in an environment. This SBG consists of: 
\begin{itemize}
	\item environment state space $S$ with fully observable kinodynamic states $s_i$ of the participants.
	\item $N$ traffic participants; for each participant $i \in N$:
		\begin{itemize}
				\item action set \(A_i\) of motion primitives
				\item behavior type space \(\Theta_i\) modeling the driving styles, e.g.   \(\Theta_i=\{\text{"aggressive},\text{"passive"}\}\). 
				\item reward function \(\mathcal{R}_i: S \times A \times \Theta_i \rightarrow \mathbb{R}\) defining the reward after executing the joint action $a\in A$.
				\item stochastic policy \(\pi_i: \mathbb{H}\times A_i \times \Theta_i \to [0,1]\) over the sets of state-action histories $\mathbb{H}$, actions \(A_i\) and behavior types \(\Theta_i\), e.g. $\pi_i(H^t, a,\text{"aggressive"})$. 
			\end{itemize}
	\item state transition function $T: S  \times A \times S \to [0,1]$. 
	\item type distribution $\Delta: \mathbb{N}_0 \times \Theta \to [0,1]$ over the sets of participants' indices $\mathbb{N}_0$ and types $\Theta$. In the above example, it reflects the percentage of drivers showing "aggressive" or "passive" behavior. 
	\end{itemize}
Before each episode of the SBG, the type $\theta_i$ for each participant~$i$ is sampled from $\Theta_i$ with probability $\Delta(\theta_i)$. Based on the state-action history $H^t$ at time step $t$, each participant repeatedly chooses an action according to its behavior $\pi_i(H^t,a,\theta_i)$ until a terminal environment state occurs. The ego-vehicle, $i{=}0$, knows a priori the type distribution~$\Delta$ and space~$\Theta$, and the behaviors $\pi_i$; in our approach via inference during training in simulation. The episode-specific, sampled types \(\theta_i\) of the other participants are unknown.

A behavior generation algorithm must solve the presented SBG. It shall find the optimal driving policy of the ego-vehicle $\pi_0(\cdot,\cdot,\cdot)$, maximizing positive return while considering the risk of negative return due to the uncertainty about the episode-specific driving styles of other participants.

\section{Method}
We propose a risk-sensitive behavior generation approach to deal with the presented problem. It encompasses the following two steps visualized in Fig.~\ref{fig:top_level}:
\begin{enumerate}
	\item \textbf{Offline Distribution Learning:} The random return variable $R$ depending on action $a$ in environment state~$s$ is distributed according to the state-action return distribution  $Z(r|s,a)$. Using Distributional Reinforcement Learning \cite{bellemare_distributional_2017}, we learn $Z^*(r|s,a)$ in simulation for a fixed behavior type space and distribution. It encodes the optimal policy of the ego-agent for such an environment.
	\item  \textbf{Online Risk Assessment:}  We deviate from using the standard, expectation-based selection of the optimal action with  $a^*{=} \argmax_a \E_{r\sim Z^*}[R]$.  Instead, we quantify the collision risk with distortion risk metrics~\cite{majumdar_how_2017} applied to the learned state-action value distribution. The optimal action is then selected based on the measured risk of each action.
\end{enumerate}
Next, we depict the presented approach in detail.
\subsection{Distributional Reinforcement Learning}
Reinforcement learning finds an optimal policy for a Markov Decision Process (MDP). The Bellman equation defines the optimal Q-function
\begin{equation} \label{eq:bellman}
Q^*(s,a) = \E_{s^\prime}\left[\mathscr{r}(s,a,s^\prime) + \gamma \max_{a^\prime} Q^*(s^\prime,a^\prime)|s,a\right]
\end{equation} 
representing the expected return, taking action \(a\) in state \(s\) and from thereon following the optimal policy~\(\pi^*(s) {=}a^*{=}  \argmax_a Q^*(s,a)\).
The discount factor \(\gamma\) defines how future rewards $\mathscr{r}_t \sim \mathscr{R} $ contribute to the current state-action value. \citet{mnih_human-level_2015} introduced Deep Q-Networks (DQN) enabling Q-learning for problems with higher-dimensional, continuous state space. Double Deep Q-Networks (DDQN) \cite{van_hasselt_deep_2016} and prioritized experience replay~\cite{schaul_prioritized_2016} improved convergence and optimality of DQN.

Distributional reinforcement learning models the return $R$ as a random variable with probability distribution $Z(r|s,a)$ and the Q-value being the expected return $Q(s,a){=}\E_{r\sim Z}[R]$. \citet{bellemare_distributional_2017} introduced Deep Distributional Reinforcement Learning to learn  $Z(r|s,a)$ non-parametrically in a continuous state space. They proved that the Distributional Bellman equation
\begin{equation}
\begin{split}
Z(r|s,a) \stackrel{\small\mbox{D}}{:=} \mathcal{R}(s,a) + \gamma Z(r|s',a') \\  \quad s'\sim T(\cdot|s,a), a'\sim \pi^*(\cdot|s')
\end{split}
\end{equation}
has a unique fix point $Z^*(s,a)$ that minimizes the maximal form of the Wasserstein metric, a distance between two probability distributions. Their proposed algorithm, C51, approximately minimizes this distance to learn the return distribution $Z^*(s,a)$ from which the optimal policy is obtained greedily with $\pi^*(s) = \argmax_a \E[Z^*(s,a)]$. 

 Quantile Regression Deep Q-Learning (QRDQN) improves the performance of the C51 algorithm by truly minimizing the Wasserstein metric \cite{dabney_distributional_2017}. It approximates the inverse cumulative distribution function (c.d.f) or quantile function $F_Z^{-1}$ at discrete probabilities.


\subsection{Training Process}
We apply the QRDQN algorithm to learn the state-action distribution $Z^*(r|s,a)$ of the ego agent for a \emph{fixed} behavior type space and distribution in simulation.

Before the start of a training episode, we sample episode-specific behavior types $\theta_i$ for all other participants $i$ from the fixed environment-specific type distribution $\Delta$. The sampled types remain constant for the rest of the episode.  The simulated participants then behave according to $\pi_i(\cdot,\cdot,\theta_i)$. By seeing a multitude of episodes with different behavior types, the learning agent infers the type distribution and space, and learns a risk-neutral, optimal policy for the given SBG. After learning, $Z^*(r|s,a)$ expresses the inherent uncertainty about the actual behavior types appearing at a specific episode.

\subsection{Risk Assessment}
During execution, we quantify the risk of an action based on the learned distribution $Z^*(s,a)$ using risk metrics. Learning of the state-action distribution occurred risk-neutral with expectation-based action selection. Now, during execution of the learned behavior, we quantify the action risks with a distortion risk metric applied to the learned risk-neutral distribution.

Distortion risk metrics comply with six mathematical axioms and emerged from the field of finance. Their application as risk metric in robotics is discussed by \citet{majumdar_how_2017}. Sequential decision making may be temporally inconsistent when risk evaluation is not applied already during training \cite{ruszczynski_risk-averse_2010}. However, the advantage of assessing risk based on the risk-neutral distribution is that the most suitable risk estimator and its parameters could be adapted online to the encountered traffic scene. 

\begin{figure}[t]	
	\vspace{1mm}																																								
	\def\svgwidth{\columnwidth}
	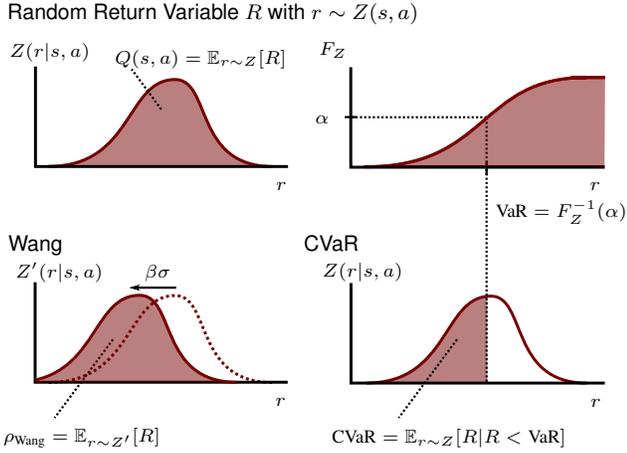 \vspace{-6mm}
	\caption{Graphical representation of the calculation of Wang and CVaR risk metrics: \emph{Wang} distorts the original distribution and calculates an expectation over the resulting distribution. It shifts the mean for a normal-like distribution. \emph{CVaR} considers only returns below the Value at Risk (VaR).} \label{fig:risk_metric_calc}
	\vspace{-3mm}
\end{figure}

We evaluate two distortion risk metrics. For better readability, we denote $Z = Z^*(r|s,a)$ in the following:
\begin{itemize}
	\item \textbf{Conditional Value at Risk} (CVaR) \cite{majumdar_how_2017}: 
	\begin{equation}
	\rho_\text{CVaR}[Z] =\E_{r\sim Z}[R|R<\text{VaR}_\alpha]
	\end{equation}
	with probability parameter $\alpha$ and the value at risk $\text{VaR}_\alpha {:=} F_Z^{-1}(\alpha)$. Thus, $\alpha$ is the cumulative probability of returns smaller than $\text{VaR}_\alpha$. The $\text{CVaR}_\alpha$ is the mean over this section of the return distribution. 
	\item \textbf{Wang} \cite{wang_class_2000}: 
	Wang distorts the original cumulative distribution and uses the expectation of the resulting distribution:
	\begin{equation}
	\begin{gathered}
	 F_{Z'}= \Phi[\Phi^{-1}(F_Z)+\beta]\\ \rho_\text{Wang}[Z]=\E_{r\sim Z'}[R]
	\end{gathered}\end{equation}
	where $\Phi$ is the standard normal c.d.f and $\beta$ a real-valued parameter. For a normal distribution, this metric shifts its mean with $\mu' {=} \mu {+} \beta\sigma$.
\end{itemize}
The calculation of the risk metrics is represented graphically in Fig. \ref{fig:risk_metric_calc}.

Action selection is greedy. For each action $a_i$, we calculate its expected return under the risk metric and select with
	\begin{equation} 
 a^*(s_t) =	\argmax_{a_i \in \mathcal{A}} \rho_\text{xxx}(Z(r|s_t,a_i)). 
	\end{equation}
the optimal, risk-sensitive action $a^*$ in state $s_t$.

\section{Experiment Setup}

\begin{figure}[b]
	\vspace{-6mm}
	\centering
	\subfloat[Turn right x 2]{\includegraphics[width=0.45\columnwidth]{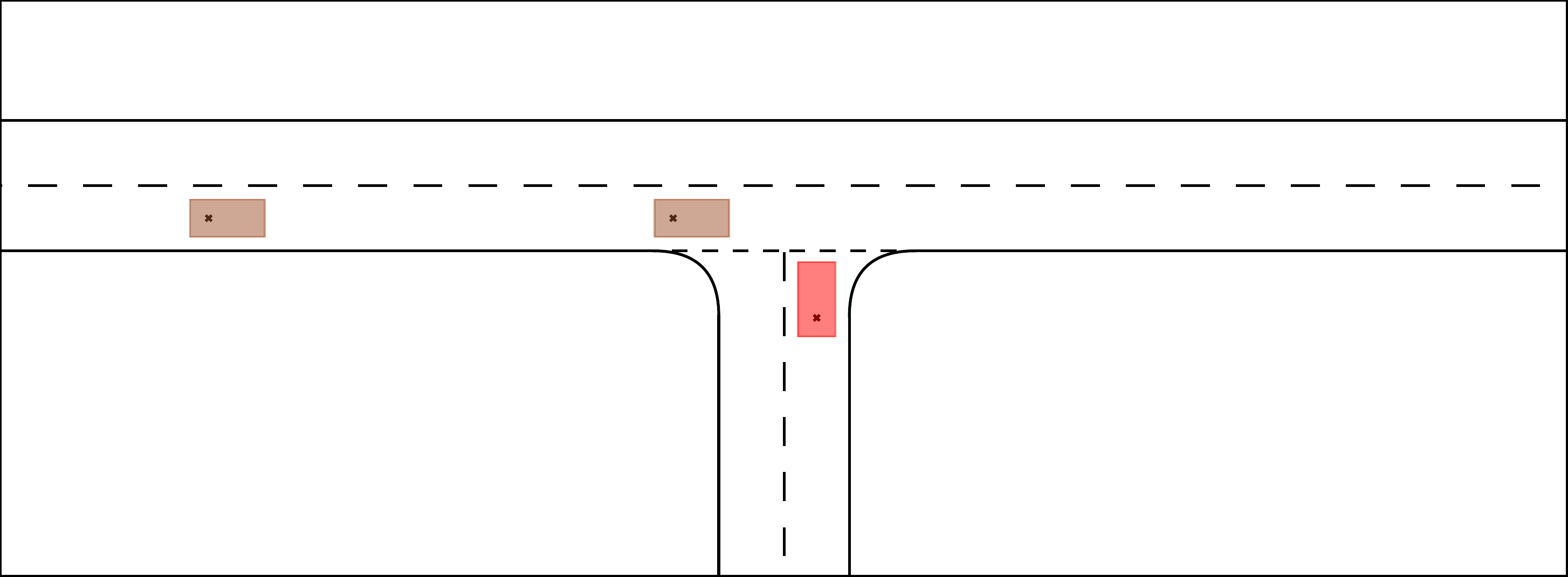}\label{fig:TurnRight} } 
	\subfloat[Turn left  x 2]{\includegraphics[width=0.45\columnwidth]{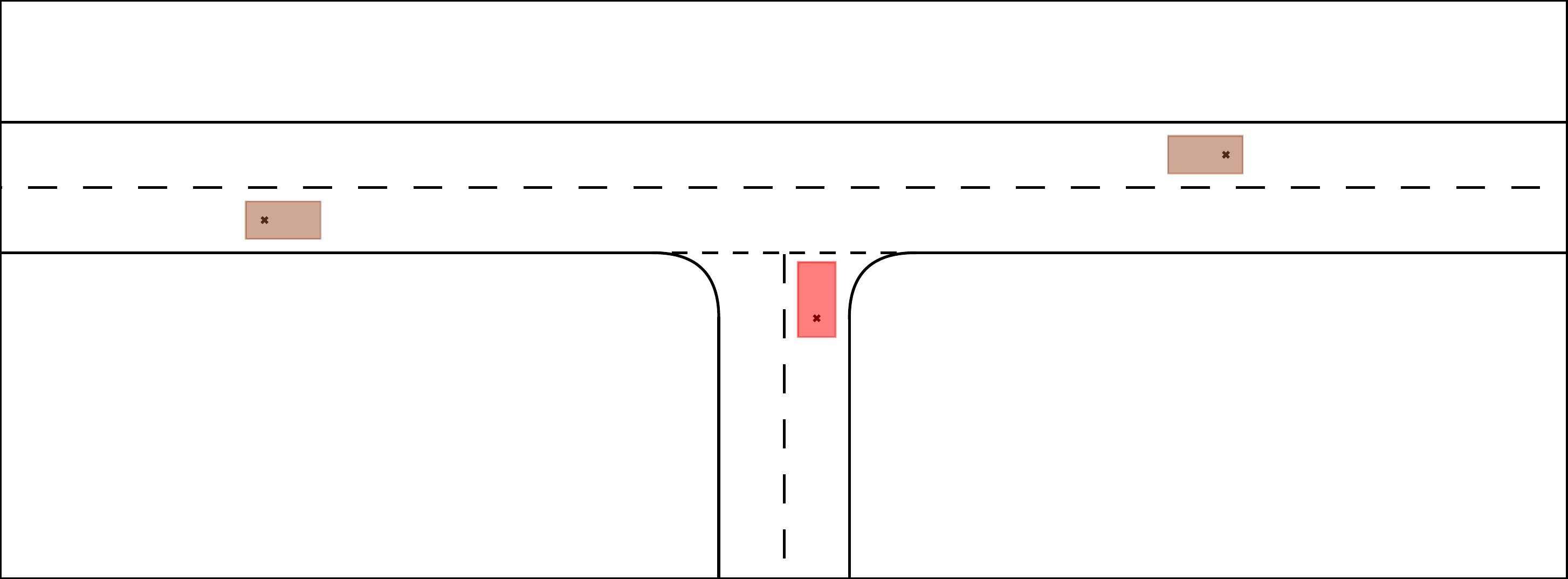}}\label{fig:TurnLeft} \\
	\subfloat[Turn left x 4]{\includegraphics[width=0.45\columnwidth]{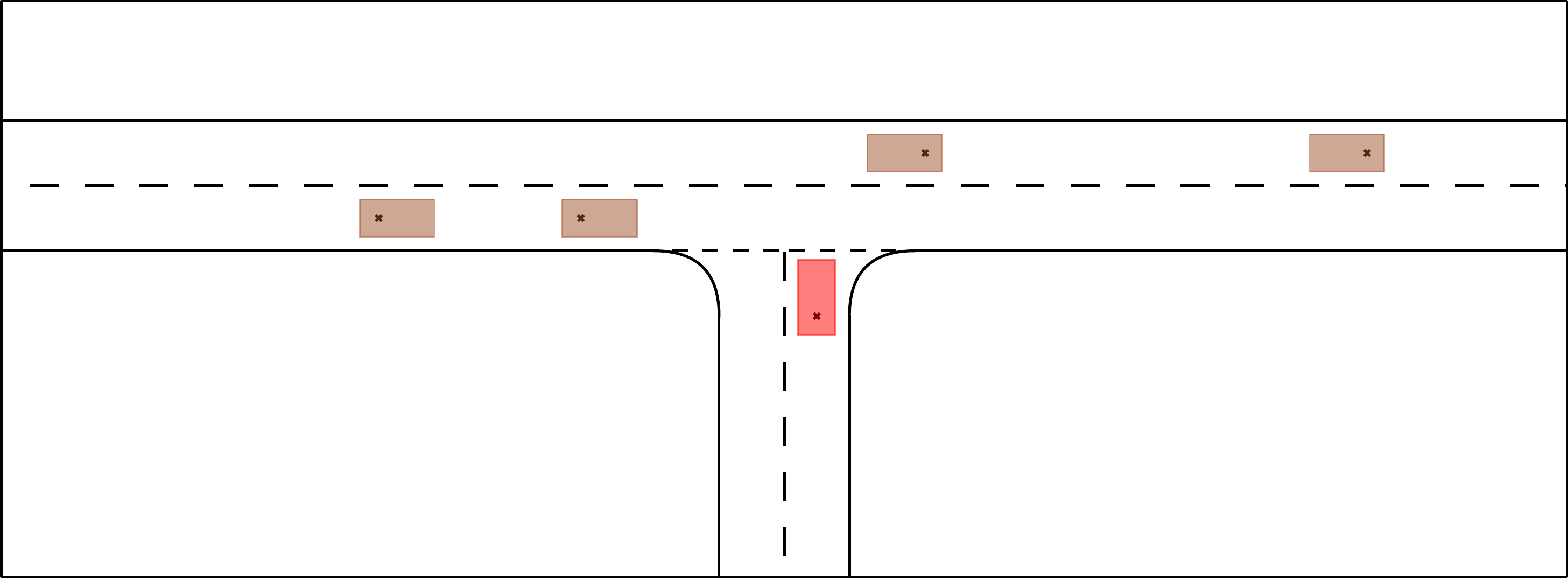} \label{fig:ComplexTurnLeft} } 
	\subfloat[Turn right platoon]{\includegraphics[width=0.45\columnwidth]{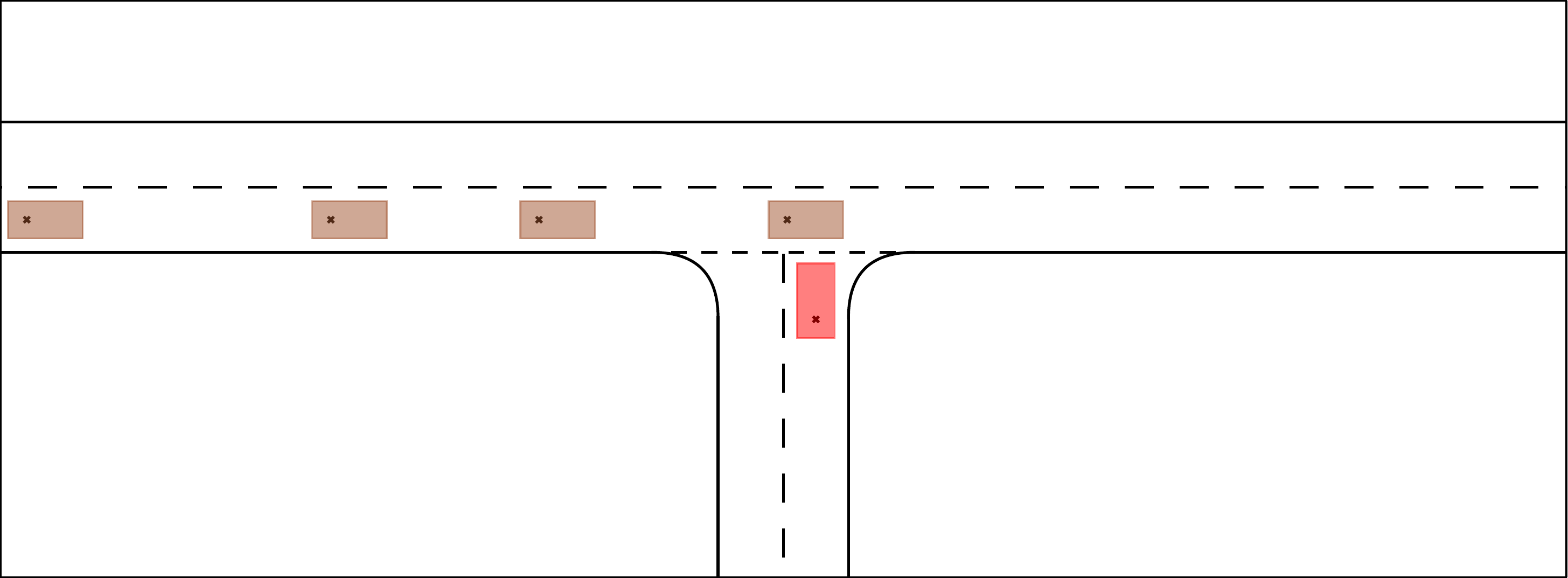}	\label{fig:PlatoonTurnRight} }
	\caption{Intersection scenarios considered in the evaluation.} \label{fig:scenarios}
	\vspace{-2mm}
\end{figure}

We evaluate our approach on four turning scenarios in the T-intersection given in Fig.~\ref{fig:scenarios}. Below, we describe the experiment setup in detail.  

\subsection{Scenario}
We consider four turning scenarios, with either left or right turn and a varying number of participants. The other participants have right of way, but react to the ego-vehicle.
At the beginning of an episode, the ego-vehicle starts at the same point of the intersection with zero initial velocity. It succeeds when reaching the end of the turning lane without collision. We limit the velocity for all participants to 54~km/h and the maximum acceleration to $\SI{5}{\metre\per\square\second}$ and $\SI{-4}{\metre\per\square\second}$.  The time step of the simulation is $\SI{200}{\milli\second}$.  

\subsection{Behavior Modeling}
We define two deterministic driving styles "passive" and "aggressive". Stochastic behavior types are planned in future work. The corresponding policies $\pi(\cdot,\cdot,\text{"passive"})$ and $\pi(\cdot,\cdot,\text{"agressive"})$ use an Intelligent Driver Model (IDM) that reacts also to turning vehicles. $\pi(\cdot,\cdot,\text{"agressive"})$ accelerates to the desired velocity and keeps the gap to other IDM vehicles. But, it does not react and brake, if the ego-vehicle occupies the lane.

As we want to evaluate performance with and without behavioral uncertainty, we consider two type definitions:
\begin{itemize}
	\item \textbf{Single:} All the other drivers behave aggressively, thus $\Theta_\text{single}{=}\{\text{"aggressive"}\}$ with $\Delta_\text{single}(\text{"aggressive"}){=}1.0$.
    \item \textbf{Mixed:} Drivers act with equal percentage passively or aggressively, thus $\Theta_\text{mixed}{=}\{\text{"passive"},\text{"aggressive"}\}$ with uniform type distribution $\Delta_\text{mixed}(\text{"passive"}){=}0.5$ and $\Delta_\text{mixed}(\text{"agressive"}){=}0.5$. 
\end{itemize}
Before an episode, a single type is sampled from the selected type distribution. It is then used by all other participants.  

\subsection{Deep Reinforcement Learning}
We train DQN and QRDQN agents separately for each scenario for both type definitions "single" and "mixed". We employ the standard DQN \cite{mnih_human-level_2015} and QRDQN \cite{dabney_distributional_2017} architectures with fully connected layers ($4\times300$~ReLUs), outputting a single value for each action, respectively $N{=}200$ quantiles for each action. The input consists of the concatenated observations of all participants. Observation and action space are given below. We use prioritized experience replay~\cite{schaul_prioritized_2016} and Double DQN~\cite{van_hasselt_deep_2016} for both DQN and QRDQN.

Rewards are defined for the ego-agent. The other participants do not adhere to reward maximization. They are fully controlled by their policies $\pi$. The ego-agent receives a positive reward $\mathcal{R}_{goal}{=}100$ for reaching the goal, and \mbox{$\mathcal{R}_{collision}{=}-1000$} for collisions. Every action costs additionally $\mathcal{R}_{step}=-5$. The discount factor $\gamma$ is set to 0.95.

\subsection{Training \& Test Data}
For each combination of scenario and type definition, we define a fixed training and test data set consisting of $\num{100000}$ episode definitions, respectively. Each episode definition contains the environment state, specifying the initial kinodynamic vehicle states of all participants at the beginning of the episode, and the applied behavior type. The distribution of behavior types in the data set complies with the selected type distribution $\Delta_\text{single}$ or $\Delta_\text{mixed}$. 

\begin{figure}
	\centering
		\vspace{2mm}
	\subfloat[Small gap]{
		\includegraphics[width=0.28\columnwidth]{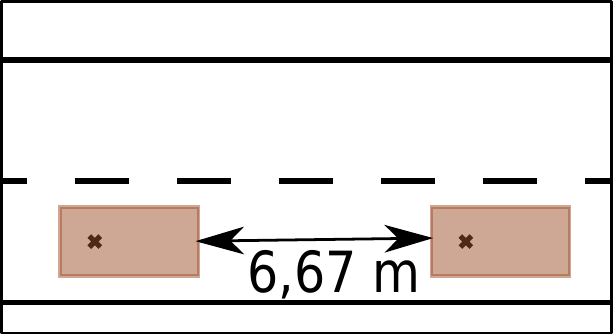} }
		\label{fig:SmallGap}  
\subfloat[Intermediate gap]{
		\includegraphics[width=0.4\columnwidth]{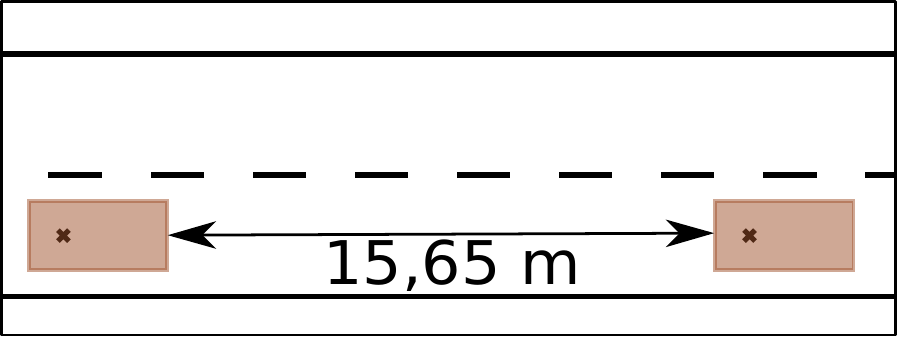} } 
		\label{fig:IntermediateGap} \\[1ex]
\subfloat[Large gap]{
		\includegraphics[width=0.38\columnwidth]{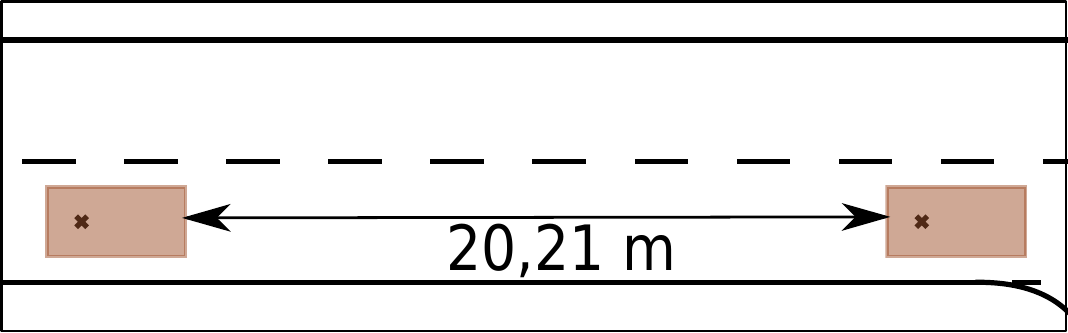} }
		\label{fig:LargeGap}
	\caption[Different gap sizes in the training data.]{Different gap sizes in the training and test data sets.}
	\label{fig:GapSize}
		\vspace{-4mm}
\end{figure}

To improve generalization, we vary the number of participants up to the maximum count of the scenario and shuffle the order of vehicle state positions in the concatenated observation space. The other participants have a random velocity between $\SI{29}{\kilo\metre\per\hour}$ and $\SI{36}{\kilo\metre\per\hour}$. Further, the data sets contain different initial gap sizes between two vehicles in the same lane as depicted in Fig.~\ref{fig:GapSize}.

\subsection{Evaluation Metrics and Significance Testing}
\label{sec:eval_metrics}

The following metrics are used for evaluation:
\begin{itemize}
	\item \textbf{Success/Collision rate} [\%]: percentage of runs the ego-vehicle reached the end of the turning lane/collided. 
	\item \textbf{Max. time rate} [\%]: percentage of runs exceeding maximum allowed crossing time (\SI{14}{\second}). 
	\item \textbf{Crossing time} [s]: time to reach the goal averaged over successful runs.
\end{itemize}
We use a fixed set of \num{10000} test runs per approach and scenario to calculate the metrics, employing the best performing training checkpoint after a fixed number of training steps.

To check for significance in performance differences, we use paired statistical tests in which the test set index is the independent variable. We perform for the binomial variables (success, collision, max. time) a Cochran's Q test followed by pair-wise McNemar tests. For the crossing time, we use a repeated measures ANOVA followed by pair-wise dependent t-tests. Confidence level is 0.95 with Bonferroni correction for the pair-wise tests.

\subsection{Action Space} \label{sec:ac_st_benchmark}
We use only longitudinal actions \(A_\text{ego} {=} (-3,0, 2, 5) \) in $\si{\metre\per\square\second}$ along predefined left or right turning paths to facilitate an empirical analysis of the benefits of our method. A generally applicable driving policy including lateral actions is deferred to future work. The other participants have a continuous action space defined by their behavior model.

\subsection{Observation Space} \label{sec:ac_st_benchmark}
To find an appropriate observation space of the intersection scenario applied as input to DQN and QRDQN, we benchmarked different observation spaces. We chose a right turn scenario with a single other participant and type distribution $\Theta_\text{single}$, and compared the success rates of a DQN agent for the following observation spaces: 

\begin{enumerate}
	\item Cartesian coordinates and velocity
	\begin{equation*}
	(x_\text{ego}, y_\text{ego}, v_\text{ego}, x_1, {y}_1, {v}_1, \ldots,  {x}_k, {y}_k, {v}_k) 
	\label{Eq:State0}
	\end{equation*}
	
	\item Cartesian coordinates, velocity and binary value
	\begin{equation*}
	( {x}_\text{ego} , {y}_\text{ego} ,  {v}_\text{ego} ,  {x}_1, {y}_1 ,  {v}_1 ,  c_1, \ldots,  {x}_k ,  {y}_k,  {v}_k ,  c_k )
	\label{Eq:State1}
	\end{equation*}
	
	\item Relative features to ego-vehicle
	\begin{equation*}
	\begin{split}
	( \Delta {x}_1 , \Delta {y}_1 ,\Delta {v}_{x,1} ,\Delta {v}_{y,1} ,\ldots  ,\\ \Delta {x}_k , \Delta {y}_k ,  \Delta {v}_{x,k} ,  \Delta {v}_{y,k} )
	\end{split}
	\label{Eq:State2}
	\end{equation*}
	
	\item Relative features and ego-vehicle state
	\begin{equation*}
	\begin{split}
	( {x}_\text{ego} , {y}_\text{ego},  {v}_\text{ego} , \Delta {x}_1 ,  \Delta {y}_1 ,  \Delta {v}_{x,1} ,  \Delta {v}_{y,1} ,  \ldots, \\ \Delta {x}_k , \Delta {y}_k ,  \Delta {v}_{x,k} , \Delta {v}_{y,k} ) \end{split}
	\end{equation*}
	
	\item Distance, orientation, velocity and TTC (inspired by~\cite{isele_navigating_2017})
	\begin{equation*}
	\begin{split}
	({x}_\text{ego} , {y}_\text{ego},  {v}_\text{ego},  {d}_{1},  \Delta {\phi}_1,  {v}_1,  \text{TTC}_{1}, \ldots, \\ {d}_{k}, \Delta {\phi}_k, {v}_k, \text{TTC}_{k}) \end{split}
	\label{Eq:State4}
	\end{equation*}
	
	\item Distance, signed velocity and lane
	\begin{equation*}
	({d}_{1}, {v}_{\phi, 1}, \text{lane}_1,\, \ldots, {d}_{k}, {v}_{\phi, k}, \text{lane}_k )
	\label{Eq:State5}
	\end{equation*}
\end{enumerate}

All real-valued numbers are normalized to the range \mbox{\num{-1} to \num{1}}. The $\Delta$ sign denotes the value difference, $d_k$ the Euclidean distance and $\text{TTC}_k$ is the Time-To-Collision between vehicle $k$ and the ego-vehicle.  The binary value $c$ is one, if the vehicle is present in the scene, otherwise, it is zero. The signed velocity $v_{\phi, k}$ is positive when driving from left to right or bottom to top, and negative when driving right to left. The lane index $\text{lane}_k$ can take a value between one and four to indicate the position on one of the four available lanes.
If the number of vehicles is lower than the scenario maximum, we calculate missing features based on a vehicle with a zeroed state, not interfering with the drivable space of the intersection, $\text{TTC}{=} \num{-1}$ and $\text{lane}{=}0$. 

Table~\ref{tab:StateRepresentationBenchmark} compares the results, in this preliminary evaluation, without significance testing. The TTC-based representation lead to an aggressive driving behavior with high collision rate. Interestingly, sparse observation spaces such as $2)$ and $6)$ achieved an acceptable overall performance. However, relative state information in combination with the ego-vehicle state (4) outperformed the other representations in terms of collision rate and achieved a medium crossing time. Thus, we decided to use this representation in the  evaluation.

\begin{table}[t]
	\centering
	\vspace{2mm}
	\caption{Results of the observation space benchmark.}
	\scriptsize
	\begin{tabular}{p{3.6cm}cccc}
		\toprule
		Observation  & Collision& Max. Time  & Crossing   \\
		Representation&  Rate [\%]  & Rate	 [\%]	&  Time [s]	\\
		\midrule		
		1) $x, y, v$ & 9.8 &0.0 & 4.53  \\
		\midrule
		2) $x, y, v, c$  & 2.6 &0.0 & 4.70  \\
		\midrule
		3) $\Delta x, \Delta y, \Delta v_x \Delta v_y$ & 23.8 &0.0 & 3.31 \\
		\midrule
		4) $\Delta x, \Delta y, \Delta v_x \Delta v_y, \text{state}_\text{ego}$ & 1.4 &0.0 & 4.06  \\
		\midrule
		5) $d, \Delta \phi, v, \text{TTC}$ & 31.9&0.0 & 3.24 \\
		\midrule
		6) $d, v_\phi, \text{lane}$ & 2.6 &0.0 & 4.79 \\
		\bottomrule
	\end{tabular}	
	\label{tab:StateRepresentationBenchmark}
	\vspace{-2mm}
\end{table}

\section{Evaluation}
In our final evaluation, we compare the performance of the DQN baseline \cite{wolf_adaptive_2018, mirchevska_high-level_2018, isele_navigating_2017}  with QRDQN, and QRDQN with risk-sensitive policy evaluation in environments with single and mixed behavior type definitions. 

\subsection{Exemplary Training Results}

\begin{figure}[t]
	\vspace{2mm}
	\centering
	\includegraphics{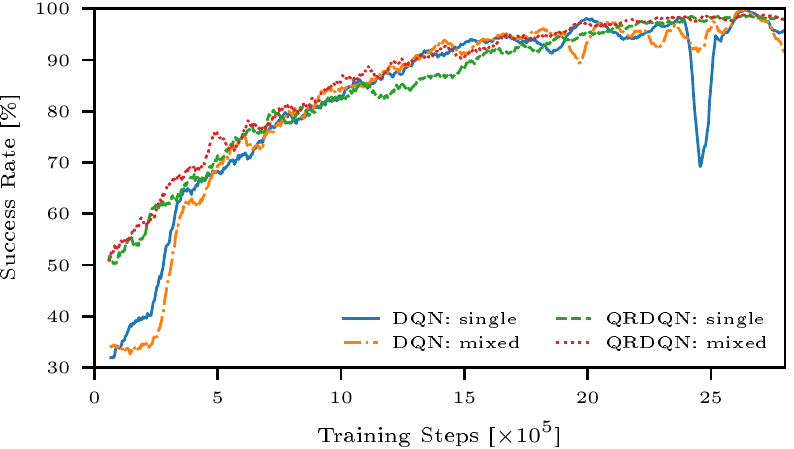}  
	\vspace{-2mm}
	\caption{Success rates during training of DQN and QRDQN for the single and mixed behavior type space in the "Turn left x 2" scenario.}
		\label{fig:training_results} 
		\vspace{-3mm}
\end{figure}

Fig. \ref{fig:training_results} compares the training data success rates of DQN and QRDQN over the course of training, exemplarily for the "Turn left x 2" scenario with single or mixed behavior type space. QRDQN converged smoothly in both the single and mixed setting. In contrast, DQN fluctuated strongly from the \num{20e5} episode on in the mixed setting. As expected, QRDQN showed thus improved stability in the training process for stochastic environments.

\subsection{Risk Metric Parameterization}
In a preliminary study, we coarsely evaluated the influence of the risk metric parameters $\alpha$ of CVaR and $\beta$ of Wang using the trained QRDQN agents. We considered the average performance over all scenarios in the training data set. We discovered that Wang was very sensitive to parameter changes and started to yield conservative driving behavior for $\beta<-0.4$. In contrast, CVaR was more robust to parameter changes. For the following evaluation, we set $\alpha=0.7$~and~$\beta = -0.2$.

\subsection{Quantitative Analysis}
First, we qualitatively compare the different approaches. We highlight in bold the best/worst result of a group, if the group test and all pair-wise tests within the group were significant as described in~\mbox{Sec.~\ref{sec:eval_metrics}}.

\begin{table}[b]
	\caption{Comparison of algorithms averaged over all scenarios.} \label{tab:algorithm_comparison}
	\centering
	\scriptsize
	\begin{tabular}{llllp{1cm}}
\toprule
      &       & \% Collisions & \% Max. Time & Crossing Time [s] \\
$\Theta_\text{others}$ & Algorithm &               &              &                   \\
\midrule
\multirow{3}{*}{single} & QRDQN + CVaR &         \textbf{1.18} &         0.00 &              5.43 \\
      & DQN &          3.09 &         0.00 &              6.12 \\
      & QRDQN &          2.10 &         0.00 &             \textbf{5.27} \\
\cmidrule{1-5}
\multirow{3}{*}{mixed} & QRDQN + CVaR &          \textbf{0.70} &         0.00 &              6.19 \\
      & DQN &         6.98 &        \textbf{24.75} &              7.45 \\
      & QRDQN &          1.68 &         0.00 &              \textbf{5.16} \\
\bottomrule
\end{tabular}

\end{table}

\begin{table}[t]
	\vspace{2mm}
	\caption{Comparison of risk metrics in the "mixed" type space.} \label{tab:scenario_comparison} 
	\centering
	\scriptsize
	\begin{tabular}{llllp{{1cm}}}
\toprule
           &      & \% Collisions & \% Max. Time & Crossing Time [s] \\
Scenario & Risk Measure &               &              &                   \\
\midrule
\multirow{3}{*}{left x 4} & CVaR &          1.08 &         0.00 &              5.43 \\
           & None &         \textbf{2.06} &         0.00 &              \textbf{5.34} \\
           & Wang &          0.88 &         0.00 &              5.46 \\
\cmidrule{1-5}
\multirow{3}{*}{right platoon} & CVaR &          0.00 &         0.00 &             \textbf{10.83} \\
           & None &          \textbf{1.15} &         0.00 &              6.98 \\
           & Wang &          - &       \textbf{100.00} &             - \\
\cmidrule{1-5}
\multirow{3}{*}{left x 2} & CVaR &          0.08 &         0.00 &              4.84 \\
           & None &          \textbf{0.61} &         0.00 &              \textbf{4.76} \\
           & Wang &          0.09 &         0.00 &              4.88 \\
\cmidrule{1-5}
\multirow{3}{*}{right x 2} & CVaR &          1.64 &         0.00 &              3.67 \\
           & None &          \textbf{2.90} &         0.00 &              \textbf{3.57} \\
           & Wang &          1.59 &         0.00 &              3.70 \\
\bottomrule
\end{tabular}

	\vspace{-5mm}
\end{table}

\begin{figure*}[t]
	\centering
	\vspace{2mm}
	\subfloat[{Learned return distributions $Z$ and modified return distributions $Z_\text{CVaR}$ with returns pruned above the VaR at specific time points of the scenario. An arrow indicates the optimal action obtained with either $a^*=\argmax_a \E[Z]$ or $a^*=\argmax_a \E[Z_\text{CVaR}]$. }\label{fig:final_result_time_plot}
	] 
	{	\includegraphics{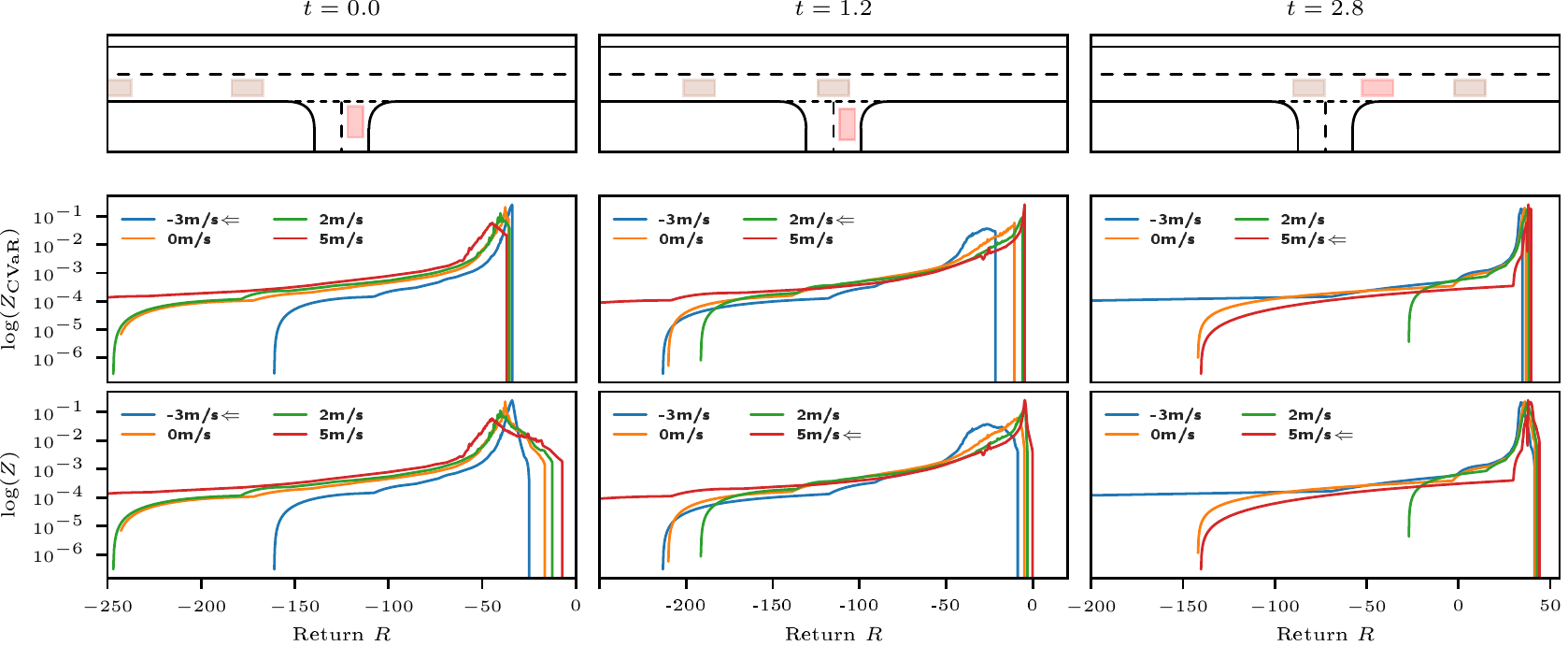}} \\
	\subfloat[Longitudinal position $s$ and velocity $v$ of the ego-agent over the course of the scenario. \label{fig:final_result_vel_plot}]  {\includegraphics{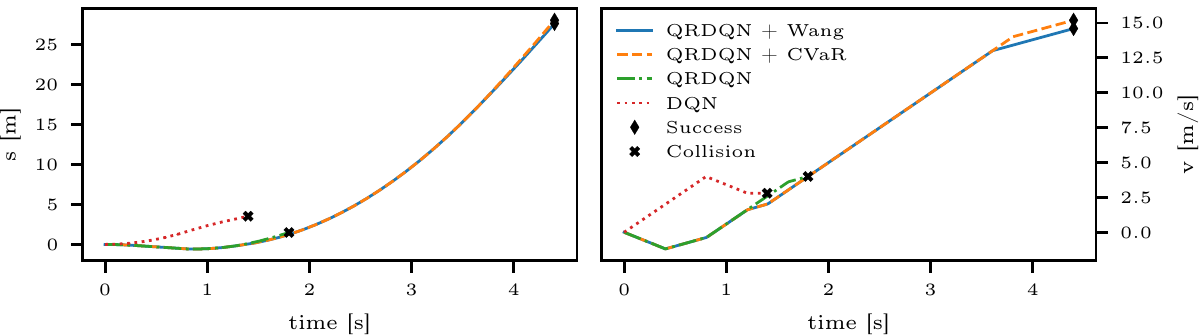} }
	\caption{For a "Turn right x 2" episode with passive behavior of other participants, we show the evaluation of the scene and the corresponding distributions at specific timepoints and compare velocity and longitudinal position for all algorithms. Risk-sensitive action selection with CVaR yields a more moderate acceleration at time point $t=\SI{1.2}{\second}$ in comparison to the learned optimal action of QRDQN choosing highest acceleration. This subtle difference made the risk-sensitive approach successfully complete this episode whereas DQN and QRDQN 
		failed.}
\vspace{-4mm}
\end{figure*}

First, we discuss the advantage of Distributional Reinforcement Learning over standard Deep Q-Learning. Table \ref{tab:algorithm_comparison} depicts the performance of these two algorithms for "single" and "mixed" behavior type space averaged over \emph{all scenarios}. For $\Theta_\text{single}$, QRDQN achieved a slightly higher success rate than standard DQN. However, for $\Theta_\text{mixed}$, when uncertainty about the behavior of others was given, QRDQN reduced collisions by 5\%. There, a local minima in the learned policy of DQN led to a large max. time rate. The crossing time decreased in the "single" and "mixed" case with QRDQN. These results underline the benefits of learning state-action \emph{distributions} $Z(s,a)$ in uncertain environments. State-action \emph{values} $Q(s,a)$ do not dissolve the subtle return nuances of such domains. A detailed, more general discussion of the benefits of the distributional approach is given in~\cite{bellemare_distributional_2017}.

Yet, a collision rate of 1.68\% remained when using QRDQN with $\Theta_\text{mixed}$. Table  \ref{tab:algorithm_comparison} provides the results for QRDQN combined with the best performing risk measure CVaR. Applying risk assessment during online planning outperformed the QRDQN approach \emph{significantly}, \emph{halving} the collision rate from 1.68\% to 0.7\%. The crossing time increased slightly due to conservative driving in the "Turn right platoon" scenario. Risk assessment led in that case to longer waiting times at the entrance of the intersection.  When no inherent uncertainty was present ($\Theta_\text{single}$), the learned distributions represent only model uncertainties, e.g. arising due to insufficient exploration or loss minimization. Risk assessment was still beneficial in that case. 

Next, we compared the CVaR and Wang risk measures. For the different scenarios, the results in the "mixed" setting are depicted in Table \ref{tab:scenario_comparison}. Pair-wise significance was found against the QRDQN approach (risk measure "none"); not between CVaR and Wang.  Still, we detect a tendency: Wang reduced collisions in \emph{three} scenarios compared to QRDQN. In the "right platoon" scenario, Wang led to conservative driving, failing to cross the intersection within the maximum episode duration. In contrast, CVaR reduced collisions in \emph{all} cases. The crossing time is comparable to QRDQN. In the platoon scenario, CVaR drives conservatively too, increasing crossing time noticeably. But in contrast to Wang, it still managed to cross the intersection in all cases. 

We conclude that CVaR is a suitable metric to evaluate risk in behavior generation algorithms of autonomous vehicles. Regarding the remaining collisions, further studies should investigate the effects of different risk metric parameterizations and evaluate the influence of epistemic uncertainties on the reliability of the proposed approach.

\subsection{Qualitative Analysis}

We pick out a single episode to qualitatively examine the reasons for better performance with risk assessment. We consider a "Turn right x 2" episode with "passive" behavior of other participants. In this case, risk assessment with CVaR or Wang resulted in successful intersection crossing, and DQN and QRDQN collided.

Fig. \ref{fig:final_result_vel_plot} depicts the longitudinal position $s$ and velocity $v$ over the course of the scenario. Slight backwards movements occurred with the distributional approaches, since,  with our reward definition, the policy was optimized solely for safety. We postpone comfort constraints to later work. For specific time points in this scenario, Fig. \ref{fig:final_result_time_plot} displays the current traffic situation, and for all actions, the corresponding learned distributions $Z$ and the modified distributions $Z_\text{CVaR}$ with returns pruned above the VaR. An arrow "$\Leftarrow$" highlights the selected, optimal action, respectively.  

Overall, we see that the distributions mainly differ in the length of a tail with lower-probability negative returns and only marginally with respect to higher probability positive returns. At the beginning of the scenario at $t{=}\SI{0}{\second}$, risk assessment with CVaR did not change the optimality of the learned action. Braking remained optimal, since the other actions' long negative tails dominate their distributions even after pruning returns above the VaR. The time point $t{=}\SI{1.2}{\second}$, however, was critical for the final outcome of the episode. The learned policy of QRDQN chose highest possible acceleration, primarily as the positive return probability in its distribution outweighs the large negative tail. In contrast, considering only the return values below the VaR for decision making yielded risk-averse action selection with the optimal action being a lower acceleration value. This avoided the collision occurring with QRDQN and led to a successful completion of the scenario ($t=\SI{2.8}{\second}$). 

This example clarifies the benefits of the CVaR metric for behavior generation in uncertain environments: Due to the inherent uncertainty about a hazardous event in the environment, the distributions of riskier actions consist of a longer, low-probability tail at negative returns facing higher probability peaks at more positive returns. The decision becomes ambiguous. The CVaR risk measure decides based on the VaR which returns to prune from the distribution, strengthening the contribution of less likely negative outcomes. Overall, this removes the ambiguity of a decision, occurring with riskier actions, and yields a saver \mbox{driving policy}.

A video comparing the performance of the evaluated algorithms and risk measures at selected episodes is found under \texttt{https://youtu.be/PSDFEG5d1xg}.

\section{Conclusion and Future Work}
We proposed a two-step approach for risk-sensitive behavior generation, evaluating the risk of actions online, based on return distributions learned offline with Deep Distributional Reinforcement Learning.
We evaluated two distortion risk metrics and demonstrated
that our approach increases safety in environments with inherent uncertainty about other participants' behaviors while avoiding too conservative driving.

\citet{majumdar_how_2017} discussed the application of risk metrics, emerging from finance, to robotics. Our approach presents now  
a step forward in applying risk-sensitive behavior generation for autonomous driving.
Yet, a distributional consideration of risk in other methods, e.g search-based methods, would broaden our understanding of its benefits and challenges.

To achieve a high level of safety under inherent \emph{and} epistemic uncertainties, we plan to combine the approach with methods reducing epistemic uncertainty in the future, e.g an additional search process \cite{bernhard_experience-based_2018}.


%

\appendices


\AtNextBibliography{\small}
\printbibliography

@article{albrecht_belief_2016,
  title = {Belief and {{Truth}} in {{Hypothesised Behaviours}}},
  author = {Albrecht, Stefano V. and Crandall, Jacob W. and Ramamoorthy, Subramanian},
  year = {2016},
  month = jun,
  volume = {235},
  pages = {63--94},
  issn = {00043702},
  doi = {10.1016/j.artint.2016.02.004},
  journal = {Artificial Intelligence},
  language = {en}
}

@article{article,
  title = {Cooperative Ramp Merging System: {{Agent}}-Based Modeling and Simulation Using Game Engine},
  author = {Wang, Ziran and Wu, Guoyuan and Boriboonsomsin, Kanok and Barth, Matthew and Han, Kyungtae and Kim, BaekGyu and Tiwari, Prashant},
  year = {2019},
  month = may,
  volume = {2},
  pages = {1--14},
  doi = {10.4271/12-02-02-0008}
}

@inproceedings{bai_intention-aware_2015,
  title = {Intention-Aware Online {{POMDP}} Planning for Autonomous Driving in a Crowd},
  booktitle = {{{IEEE International Conference}} on {{Robotics}} and {{Automation}} ({{ICRA}})},
  author = {Bai, Haoyu and Cai, Shaojun and Ye, Nan and Hsu, David and Lee, Wee Sun},
  year = {2015},
  pages = {454--460},
  publisher = {{IEEE}}
}

@article{bellemare_distributional_2017,
  title = {A Distributional Perspective on Reinforcement Learning},
  author = {Bellemare, Marc G. and Dabney, Will and Munos, R{\'e}mi},
  year = {2017},
  volume = {abs/1707.06887},
  journal = {CoRR}
}

@inproceedings{bernhard_experience-based_2018,
  title = {Experience-{{Based Heuristic Search}}: {{Robust Motion Planning}} with {{Deep Q}}-{{Learning}}},
  booktitle = {21st {{International Conference}} on {{Intelligent Transportation Systems}} ({{ITSC}})},
  author = {Bernhard, Julian and Gieselmann, Robert and Esterle, Klemens and Knoll, Alois},
  year = {2018},
  publisher = {{IEEE}}
}

@inproceedings{bouton_belief_2017,
  title = {Belief State Planning for Autonomously Navigating Urban Intersections},
  booktitle = {Intelligent {{Vehicles Symposium}} ({{IV}})},
  author = {Bouton, M. and Cosgun, A. and Kochenderfer, M. J.},
  year = {2017},
  pages = {825--830},
  publisher = {{IEEE}},
  doi = {10.1109/IVS.2017.7995818}
}

@inproceedings{bouton_reinforcement_2018,
  title = {Reinforcement Learning with Probabilistic Guarantees for Autonomous Driving},
  booktitle = {Workshop on {{Safety}}, {{Risk}} and {{Uncertainty}} in {{Reinforcement Learning}}, {{Conference}} on {{Uncertainty}} in {{Artifical Intelligence}} ({{UAI}})},
  author = {Bouton, Maxime and Karlsson, Jesper and Nakhaei, Alireza and Fujimura, Kikuo and Kochenderfer, Mykel J. and Tumova, Jana},
  year = {2018}
}

@inproceedings{burger_cooperative_2018,
  title = {Cooperative {{Multiple Vehicle Trajectory Planning}} Using {{MIQP}}},
  booktitle = {21st {{International Conference}} on {{Intelligent Transportation Systems}} ({{ITSC}})},
  author = {Burger, Christoph and Lauer, Martin},
  year = {2018},
  publisher = {{IEEE}}
}

@article{dabney_distributional_2017,
  title = {Distributional {{Reinforcement Learning}} with {{Quantile Regression}}},
  author = {Dabney, Will and Rowland, Mark and Bellemare, Marc G. and Munos, R{\'e}mi},
  year = {2017},
  volume = {abs/1710.10044},
  journal = {CoRR}
}

@inproceedings{dabney_implicit_2018,
  title = {Implicit {{Quantile Networks}} for {{Distributional Reinforcement Learning}}},
  booktitle = {35th {{International Conference}} on {{Machine Learning}} ({{ICML}})},
  author = {Dabney, Will and Ostrovski, Georg and Silver, David and Munos, Remi},
  year = {2018},
  month = jul,
  volume = {80},
  pages = {1096--1105},
  publisher = {{PMLR}},
  address = {{Stockholmsm\"assan, Stockholm Sweden}},
  series = {Proceedings of {{Machine Learning Research}}}
}

@inproceedings{dilokthanakul_deep_2018,
  title = {Deep {{Reinforcement Learning}} with {{Risk}}-{{Seeking Exploration}}},
  booktitle = {From {{Animals}} to {{Animats}} 15},
  author = {Dilokthanakul, Nat and Shanahan, Murray},
  editor = {Manoonpong, Poramate and Larsen, J{\o}rgen Christian and Xiong, Xiaofeng and Hallam, John and Triesch, Jochen},
  year = {2018},
  pages = {201--211},
  publisher = {{Springer International Publishing}},
  address = {{Cham}},
  isbn = {978-3-319-97628-0}
}

@article{garcia_comprehensive_2015,
  title = {A {{Comprehensive Survey}} on {{Safe Reinforcement Learning}}},
  author = {Garc{\'i}a, Javier and Fern{\'a}ndez, Fernando},
  year = {2015},
  volume = {16},
  pages = {1437--1480},
  journal = {Journal of Machine Learning Research}
}

@inproceedings{hubmann_belief_2018,
  title = {A {{Belief State Planner}} for {{Interactive Merge Maneuvers}} in {{Congested Traffic}}},
  booktitle = {21st {{International Conference}} on {{Intelligent Transportation Systems}} ({{ITSC}})},
  author = {Hubmann, Constantin and Schulz, Jens and L{\"o}chner, Julian and Burschka, Darius},
  year = {2018},
  publisher = {{IEEE}}
}

@inproceedings{inproceedings,
  title = {Autonomous Vehicles Testing Methods Review},
  author = {Huang, WuLing and Wang, Kunfeng and Yisheng, Lv and Zhu, Fenghua},
  year = {2016},
  month = nov,
  pages = {163--168},
  doi = {10.1109/ITSC.2016.7795548}
}

@article{isele_navigating_2017,
  title = {Navigating {{Occluded Intersections}} with {{Autonomous Vehicles}} Using {{Deep Reinforcement Learning}}},
  author = {Isele, David and Rahimi, Reza and Cosgun, Akansel and Subramanian, Kaushik and Fujimura, Kikuo},
  year = {2017},
  month = may
}

@inproceedings{isele_navigating_2018,
  title = {Navigating {{Occluded Intersections}} with {{Autonomous Vehicles Using Deep Reinforcement Learning}}},
  booktitle = {International {{Conference}} on {{Robotics}} and {{Automation}} ({{ICRA}})},
  author = {Isele, David and Rahimi, Reza and Cosgun, Akansel and Subramanian, Kaushik and Fujimura, Kikuo},
  year = {2018},
  pages = {2034--2039},
  publisher = {{IEEE}}
}

@article{kurzer_decentralized_2018,
  title = {Decentralized {{Cooperative Planning}} for {{Automated Vehicles}} with {{Continuous Monte Carlo Tree Search}}},
  author = {Kurzer, Karl and Engelhorn, Florian and Z{\"o}llner, J. Marius},
  year = {2018},
  volume = {abs/1809.03200},
  journal = {CoRR}
}

@inproceedings{kurzer_decentralized_2018-1,
  title = {Decentralized {{Cooperative Planning}} for {{Automated Vehicles}} with {{Hierarchical Monte Carlo Tree Search}}},
  booktitle = {Intelligent {{Vehicles Symposium}} ({{IV}})},
  author = {Kurzer, Karl and Zhou, Chenyang and Z{\"o}llner, Johann Marius},
  year = {2018},
  month = jun,
  pages = {529--536},
  address = {{Changshu, Suzhou, China}},
  doi = {10.1109/IVS.2018.8500712},
  bibsource = {dblp computer science bibliography, https://dblp.org},
  crossref = {DBLP:conf/ivs/2018}
}

@inproceedings{lenz_tactical_2016,
  title = {Tactical Cooperative Planning for Autonomous Highway Driving Using {{Monte}}-{{Carlo Tree Search}}},
  booktitle = {Intelligent {{Vehicles Symposium}} ({{IV}})},
  author = {Lenz, David and Kessler, Tobias and Knoll, Alois},
  year = {2016},
  pages = {447--453},
  publisher = {{IEEE}}
}

@article{majumdar_how_2017,
  title = {How {{Should}} a {{Robot Assess Risk}}? {{Towards}} an {{Axiomatic Theory}} of {{Risk}} in {{Robotics}}},
  author = {Majumdar, Anirudha and Pavone, Marco},
  year = {2017},
  volume = {abs/1710.11040},
  journal = {CoRR}
}

@inproceedings{majumdar_risk-sensitive_2017,
  title = {Risk-Sensitive Inverse Reinforcement Learning via Coherent Risk Models},
  booktitle = {Robotics: {{Science}} and {{Systems}}},
  author = {Majumdar, A. and Singh, S. and Mandlekar, A. and Pavone, M.},
  year = {2017}
}

@inproceedings{menendez-romero_courtesy_2018,
  title = {Courtesy {{Behavior}} for {{Highly Automated Vehicles}} on {{Highway Interchanges}}},
  booktitle = {Intelligent {{Vehicles Symposium}} ({{IV}})},
  author = {{Men{\'e}ndez-Romero}, C. and Sezer, M. and Winkler, F. and Dornhege, C. and Burgard, W.},
  year = {2018},
  month = jun,
  pages = {943--948},
  publisher = {{IEEE}},
  issn = {1931-0587},
  doi = {10.1109/IVS.2018.8500407}
}

@inproceedings{mirchevska_high-level_2018,
  title = {High-{{Level Decision Making}} for {{Safe}} and {{Reasonable Autonomous Lane Changing Using Reinforcement Learning}}},
  booktitle = {21st {{International Conference}} on {{Intelligent Transportation Systems}} ({{ITSC}})},
  author = {Mirchevska, Branka and Pek, Christian and Werling, Moritz and Althoff, Matthias and Boedecker, Joschka},
  year = {2018},
  publisher = {{IEEE}},
  language = {en}
}

@article{mnih_human-level_2015,
  title = {Human-Level Control through Deep Reinforcement Learning},
  author = {Mnih, Volodymyr and Kavukcuoglu, Koray and Silver, David and Rusu, Andrei A. and Veness, Joel and Bellemare, Marc G. and Graves, Alex and Riedmiller, Martin and Fidjeland, Andreas K. and Ostrovski, Georg and Petersen, Stig and Beattie, Charles and Sadik, Amir and Antonoglou, Ioannis and King, Helen and Kumaran, Dharshan and Wierstra, Daan and Legg, Shane and Hassabis, Demis},
  year = {2015},
  month = feb,
  volume = {518},
  pages = {529--533},
  issn = {0028-0836, 1476-4687},
  doi = {10.1038/nature14236},
  journal = {Nature},
  number = {7540}
}

@inproceedings{mukadam_tactical_2017,
  title = {Tactical {{Decision Making}} for {{Lane Changing}} with {{Deep Reinforcement Learning}}},
  booktitle = {Conference on {{Neural Information Processing}} ({{NIPS}})},
  author = {Mukadam, Mustafa and Cosgun, Akansel and Alireza, Nakhaei and Kikuo, Fujimura},
  year = {2017}
}

@inproceedings{paxton_combining_2017,
  title = {Combining Neural Networks and Tree Search for Task and Motion Planning in Challenging Environments},
  shorttitle = {{{IROS}}},
  booktitle = {International {{Conference}} on {{Intelligent Robots}} and {{Systems}}},
  author = {Paxton, C. and Raman, V. and Hager, G. D. and Kobilarov, M.},
  year = {2017},
  month = sep,
  pages = {6059--6066},
  publisher = {{IEEE}},
  doi = {10.1109/IROS.2017.8206505}
}

@article{ruszczynski_risk-averse_2010,
  title = {Risk-Averse Dynamic Programming for {{Markov}} Decision Processes},
  author = {Ruszczy{\'n}ski, Andrzej},
  year = {2010},
  month = oct,
  volume = {125},
  pages = {235--261},
  issn = {1436-4646},
  doi = {10.1007/s10107-010-0393-3},
  day = {01},
  journal = {Mathematical Programming},
  number = {2}
}

@inproceedings{sadigh_planning_2016,
  title = {Planning for {{Autonomous Cars}} That {{Leverages Effects}} on {{Human Actions}}},
  booktitle = {Proceedings of the {{Robotics}}: {{Science}} and {{Systems Conference}} ({{RSS}})},
  author = {Sadigh, Dorsa and Sastry, Shankar and Seshia, Sanjit A. and Dragan, Anca D.},
  year = {2016},
  month = jun,
  optpages = {66--73}
}

@inproceedings{schaul_prioritized_2016,
  title = {Prioritized Experience Replay},
  booktitle = {International {{Conference}} on {{Learning Representations}} ({{ICLR}})},
  author = {Schaul, Tom and Quan, John and Antonoglou, Ioannis and Silver, David},
  year = {2016}
}

@article{shalev-shwartz_safe_2016,
  title = {Safe, {{Multi}}-{{Agent}}, {{Reinforcement Learning}} for {{Autonomous Driving}}},
  author = {{Shalev-Shwartz}, Shai and Shammah, Shaked and Shashua, Amnon},
  year = {2016},
  volume = {abs/1610.03295},
  journal = {CoRR}
}

@article{ttc,
  title = {Extended Time-to-Collision Measures for Road Traffic Safety Assessment},
  author = {Minderhoud, Michiel M. and Bovy, Piet H.L.},
  year = {2001},
  volume = {33},
  pages = {89--97},
  journal = {Accident Analysis and Prevention},
  number = {1}
}

@inproceedings{van_hasselt_deep_2016,
  title = {Deep {{Reinforcement Learning}} with {{Double Q}}-{{Learning}}},
  booktitle = {30th {{AAAI Conference}} on {{Artificial Intelligence}}},
  author = {Van Hasselt, Hado and Guez, Arthur and Silver, David},
  year = {2016},
  pages = {2094--2100},
  publisher = {{AAAI Press}},
  address = {{Phoenix, Arizona}},
  acmid = {3016191},
  numpages = {7},
  series = {{{AAAI}}'16}
}

@article{wang_class_2000,
  title = {A {{Class}} of {{Distortion Operators}} for {{Pricing Financial}} and {{Insurance Risks}}},
  author = {Wang, Shaun S.},
  year = {2000},
  volume = {67},
  pages = {15--36},
  publisher = {[American Risk and Insurance Association, Wiley]},
  issn = {00224367, 15396975},
  journal = {The Journal of Risk and Insurance},
  number = {1}
}

@inproceedings{wolf_adaptive_2018,
  title = {Adaptive {{Behavior Generation}} for {{Autonomous Driving}} Using {{Deep Reinforcement Learning}} with {{Compact Semantic States}}},
  booktitle = {2018 {{IEEE Intelligent Vehicles Symposium}} ({{IV}})},
  author = {Wolf, Peter and Kurzer, Karl and Wingert, Tobias and Kuhnt, Florian and Z{\"o}llner, Johann Marius},
  year = {2018},
  pages = {993--1000},
  publisher = {{IEEE}}
}

%


\end{document}